# Volumetric Data Exploration with Machine Learning-Aided Visualization in Neutron Science

**Yawei Hui ***

Computer Science and Mathematics Division,
Oak Ridge National Laboratory, Oak Ridge, TN 37831

**Yaohua Liu ****

Quantum Condensed Matter Division,
Oak Ridge National Laboratory, Oak Ridge, TN 37831

**Abstract.** Recent advancements in neutron and X-ray sources, instrumentation and data collection modes have significantly increased the experimental data size (which could easily contain $10^8$-$10^{10}$ data points), so that conventional volumetric visualization approaches become inefficient for both still imaging and interactive OpenGL rendition in a 3D setting. We introduce a new approach based on the unsupervised machine learning algorithm, Density-Based Spatial Clustering of Applications with Noise (DBSCAN), to efficiently analyze and visualize large volumetric datasets. Here we present two examples of analyzing and visualizing datasets from the diffuse scattering experiment of a single crystal sample and the tomographic reconstruction of a neutron scanning of a turbine blade. We found that by using the intensity as the weighting factor in the clustering process, DBSCAN becomes very effective in de-noising and feature/boundary detection, and thus enables better visualization of the hierarchical internal structures of the neutron scattering data.

**Keywords:** Scientific visualization, feature extraction, unsupervised learning and clustering, volumetric dataset.

## 1    Introduction

It has been a long-term challenge to effectively visualize 3D objects derived from large volumetric datasets in many scientific disciplines, industry domains and medical applications [1-3]. Most implemented techniques focus on the direct volume-rendering (DVR) algorithm which excels in its high sensitivity to the delicate structures of the 3D objects at the expense of computational costs. For moderately sized datasets (typically one to ten million data points) of simple density profiles, it is relatively easy to manipulate transfer functions (TF) [4, 5] used in DVR (e.g., threshold cut-off and segmented alpha ranges) so that independent features of the 3D object and the boundaries between the signal and background noise can be well determined. However, when the complexity of the internal structures or simply the sizes of the datasets increase, it enters the domain of large datasets (typically a

---
*huiy@ornl.gov, **liuyh@ornl.gov

hundred million to ten billion data points) and a simple scheme involving only TF manipulation can no longer work efficiently.

Here we propose a new visualization analysis approach that, beside the TF manipulation, takes accounts of spatial statistics of the data points. This approach enables one to explore fine structures in the sense of spatial clustering of 3D objects. As a preliminary and yet crucial step in the visualization workflow, this analysis will play the roles of noise filtering, feature extraction with boundary detection, and generating well-defined subsets of data for the final visualization.

Among several algorithms that we have tested (including kMeans, Independent Component Analysis, Principle Component Analysis, Blind Linear Unmixing, etc.), we found that the Density-Based Spatial Clustering of Applications with Noise (DBSCAN) [6] is very effective to accomplish the clustering tasks for our visualization analysis. Surprisingly, there are very limited applications of this algorithm for 3D datasets so far [7-10]. Therefore, we have explored thoroughly, for the first time, this algorithm for its ability in detecting/identifying 3D features and creating visualization from large volumetric datasets.

Two exemplary applications of our method have been presented with neutron datasets from a single crystal diffuse scattering experiment and a neutron tomography imaging reconstruction, acquired at the Spallation Neutron Source (SNS) and the High Flux Isotope Reactor (HFIR), respectively, at Oak Ridge National Laboratory (ORNL). Recent advancements in neutron sources, instrumentation and data collection modes have pushed the size of experimental datasets into the big data domain, which poses challenges for both still imaging and interactive OpenGL rendition in a 3D setting. Interestingly, our work shows that, by using the intensity as the weighting factor in the clustering process, DBSCAN enables one to spatially separate and extract interesting scattering features from the bulk data. A single feature or a combination of many of them could be chosen to create concise yet highly informative 2D projections of the 3D objects (i.e. still imaging), or to render 3D OpenGL objects interactively so that one could explore the datasets in much more details by moving, rotating and zooming in/out around them.

In following sections, we will focus on the visualization analysis in Section 2, the applications of DBSCAN on neutron datasets in Section 3, some discussions and perspectives on future work in Section 4, and a brief conclusion in Section 5.

## 2     Visualization Analysis

The goal of our visualization procedure is to explore and identify independent features for visualization and eliminate the noise at the same time, with as less user interference as possible. In this section, we will first introduce some specific characteristics of the neutron datasets used in our analysis and show the traditional DVR visualization via solely TF manipulation on these datasets. We then lay out the details of the DBSCAN-aided visualization analysis.

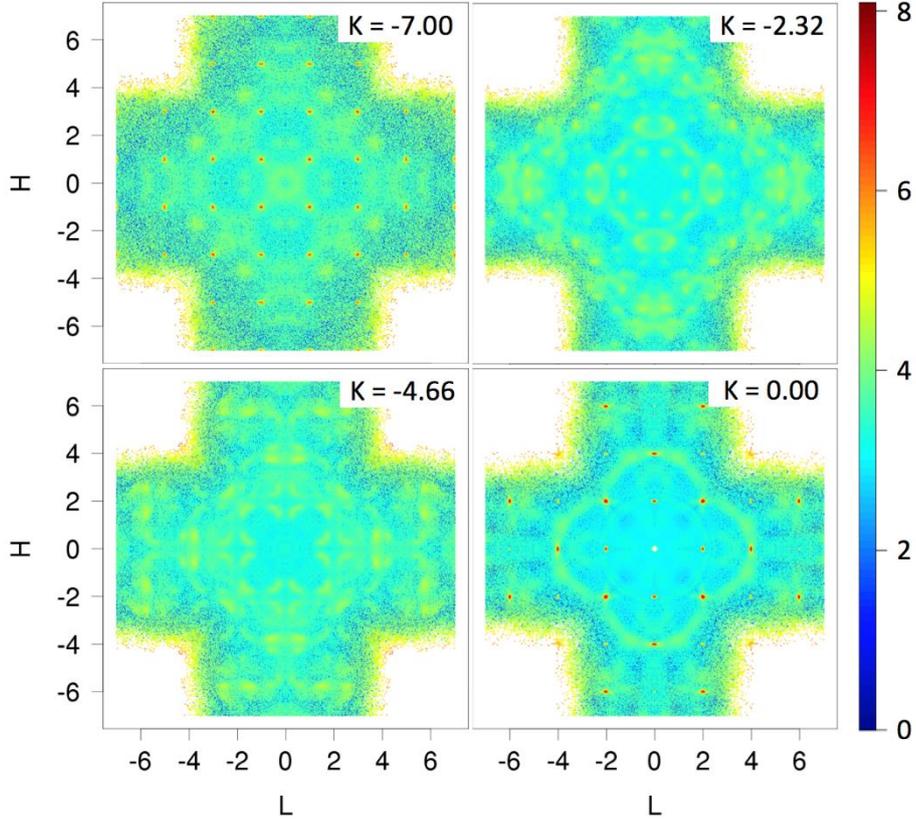

**Fig. 1.** Exemplary 2D cross-section images of the single crystal diffuse scattering data from CZO. The 2D slices were cut perpendicular to the K axis in the sample's reciprocal space with a thickness of 0.02 rlu. Data are plotted with relative scattering intensity in the logarithmic scale.

## 2.1 Characteristics of the 3D Neutron Data

The first dataset used in our analysis was collected at the elastic diffuse scattering spectrometer beamline CORELLI at SNS on a sample of single-crystal calcium-stabilized zirconia of composition $Zr_{0.85}Ca_{0.15}O_{1.85}$ (CZO hereafter). The experimental data have been reduced into the sample's reciprocal space using Mantid [11, 12]. The reduced scattering dataset is saved as a 701 x 701 x 701 3D matrix with dimensions along the H, K and L axis of the (evenly spaced) reciprocal lattice. Figure 1 shows several exemplary 2D slices cut perpendicularly to the K axis. The intensity of CORELLI data typically spans a high dynamic range (~6 orders of difference in magnitude). There exist both Bragg peaks (strong and sharp spots seen in slices K = -7 and K = 0) and diffuse scattering (broad and weak features pervasive in all slices). All "NaN" values which represent no experimental data are pre-emptively removed from the analysis. These sliced images clearly show intricate features existing in the 3D space, which are of researchers' most interest to be extracted/visualized.

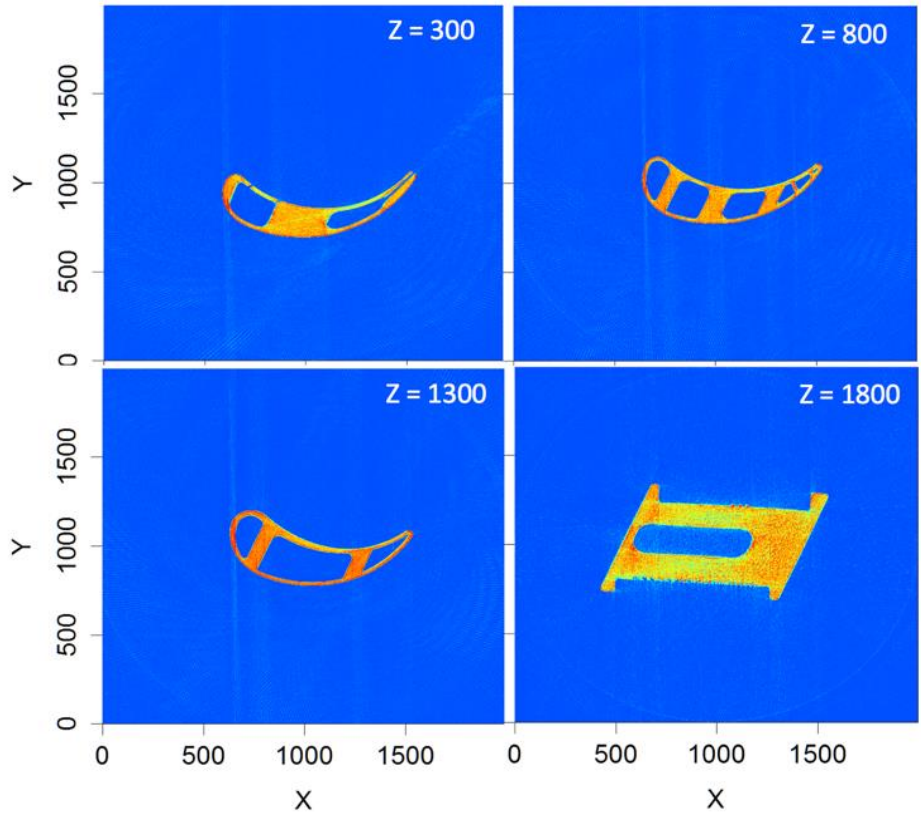

**Fig. 2.** 2D cross-section images along the Z-axis in the real space after the tomography reconstruction for an Inconel 718 turbine blade, imaged at the CG-1D cold neutron imaging prototype facility at HFIR of ORNL. The background noise can be clearly seen in these images.

The second dataset was created from a tomographical reconstruction process on an Inconel 718 turbine blade (Turbine hereafter), imaged at the CG-1D cold neutron imaging prototype facility at HFIR of ORNL [2]. After the reconstruction, the Turbine dataset was saved in a 1997 x 1997 x 1997 matrix with dimensions along the X, Y, Z axis of the 3D real space. Figure 2 shows selected 2D sliced images at different Z-positions. The noise can be seen in these images as both bulky background and filaments which are the relics of the tomography reconstruction algorithm. An efficient way to filter out these background noise before the 3D visualization is needed.

In both cases, the intensity of each data point has certain physical meaning. For the Turbine data, the intensity reflects the amplitude of the interaction potential between neutrons and the sample in the real space; while for the single crystal diffuse scattering data, the intensity is the 3D spatial Fourier transformation of the neutron-sample interaction potential. Strong localization of the measured intensities corresponds to certain physical properties of the objects under investigation and it inspires us to exploit their spatial correlations in our visualization analysis.

## 2.2    Traditional DVR with TF Manipulation

A scrutiny on the intensity profile (e.g. a histogram of the intensity) along with intuition gained from 2D cross-section images (Figure 1 and 2) reveals that many interesting structures/features are often mingled together and mixed with the vast background of noise, therefore investigating the intensity profile alone is not effective for feature detection and extraction, as illustrated in Figure 3.

We show in Figure 3(a) the histogram of intensity of the CZO dataset with several local extrema identified and in Figure 3(b) a scatterplot of the 3D still image. For clarity, the visualization space is limited to a cuboid with dimensions of (301, 501, 301) along the axis of H, K and L in the reciprocal space.

The first extreme marked at "CUSP" in the histogram reflects the random fluctuation of the background noises in the "empty" reciprocal space where the scattering signal from the sample and instrument is vanishingly small. Therefore, we set the cut-off intensity at "CUSP", which marks and removes 0.5% of the data points as "noise". For the rest of the "signal" data points, the dynamic range in their intensities spans more than 6 orders of magnitude and the total number of data points remains ~45 millions (301 x 501 x 301) in total. To avoid the overlapping problem in the traditional 3D scatterplot, we manipulate the TF by using a non-uniform "alpha" or transparency when plotting (otherwise, the scatterplot will simply manifest as a solid colored block). On the other hand, to visualize both the weak diffuse features and the strong Bragg peaks which reside at the opposite ends in the intensity profile, we divide the range of alpha at the value "THRESHOLD" (chosen to be $50 \times TOP$ after many trials) into two segments. The first one covers the weak signals in the intensity range [CUSP, THRESHOLD] with the alpha values changing linearly from 0 to 1; the second keeps a constant alpha value (=1) for all points with intensities above the THRESHOLD. Practically, it is of a trial-and-error matter to choose a proper value for THRESHOLD and it requires good understanding of the datasets. Even though some structured 3D diffuse scattering features show up in Figure 3(b), the patterns are overall vague and cloudy with noises, making it very difficult to characterize the morphological features.

## 2.3    DBSCAN-Assisted DVR

In our new approach, before employing traditional DVR for 3D visualization, we reduce the data via the unsupervised clustering algorithm DBSCAN to remove noise points by default and to facilitate the feature extraction and object-boundary detection.

The general application of DBSCAN takes two parameters in the clustering procedure: $\varepsilon$ - the maximum distance between two points for them to reside in the same neighborhood; and minPts - the minimum number of points required to form a dense region. The distance between two points is usually defined in the Euclidean metric. In the neutron datasets discussed here, coordinates of the data points are taken as the uniformly distributed voxel indices which scale linearly with the physical positions of data points in either the reciprocal space for the CZO dataset, or the real space for the Turbine. The calculation of the parameter $\varepsilon$ would then be greatly simplified. For a Cartesian coordinate system, if considering only the smallest "$\varepsilon$- neighborhood" of an arbitrary point, we can set ε between [1, $\sqrt{2}$] which will include only the six first nearest neighbors. To expand the ε-neighborhood, we can choose the value of ε in [$\sqrt{2}$, $\sqrt{3}$] to include the twelve second nearest neighbors, and so on. In the following visualization analysis, we keep ε fixed at 1.7 so that only the 18 nearest neighbors (1st and 2nd) are taken into the calculation of the density for local clustering.

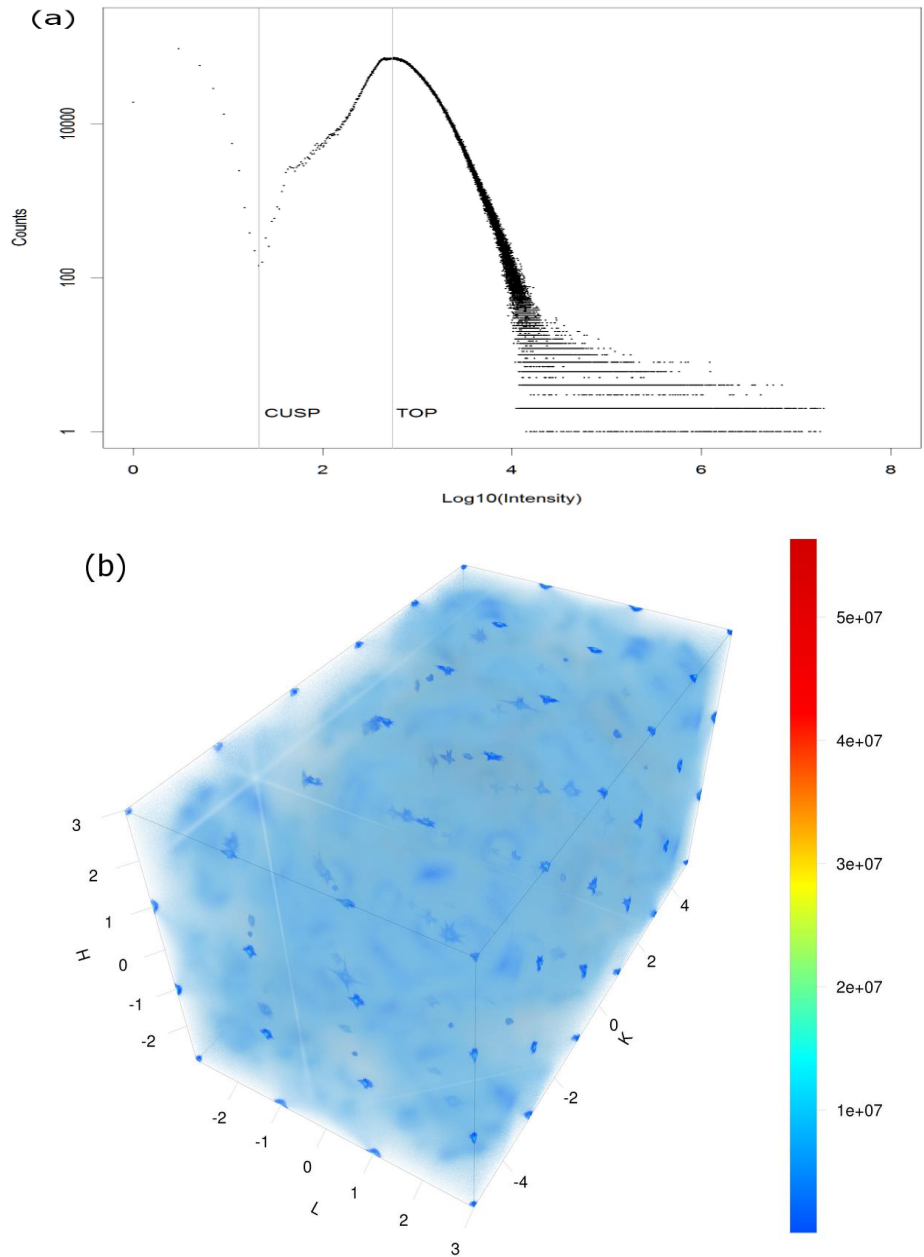

**Fig. 3.** (a) Intensity histogram of the CZO dataset in a cuboid with dimensions (301, 501, 301) along (H, K, L) axis. (b) 3D still image of the CZO dataset obtained solely by TF segmentation. Data are shown on the linear scale of intensities with noises removed below "CUSP" and a segmented TF divided at THRESHOLD = 50×TOP.

The most critical adaptation to apply DBSCAN on our neutron datasets is to use the intensity as a measure of weight in calculating the second DBSCAN parameter – minPts. As mentioned above, the intensity of neutron scattering data is of physical significance, however the traditional DVR algorithm doesn't take it into account when designing the TF. It's obvious that the diffuse scattering features shown in both 2D (Figure 1) and 3D images (Figure 3(b)) are as much spatially correlated as photometrically. To utilize both information, i.e., the intensity and spatial location, we dictate the algorithm to calculate minPts with varying weights so that for each data point, its contribution to weighted-minPts is proportional to its intensity. By doing so, the DBSCAN algorithm becomes very effective in de-noising and feature/boundary detection for neutron scattering data. For example, for the CZO dataset, with a proper weighted-minPts value, one can detect both the Bragg peaks (sharp spots with a few high intensity points) and the diffuse scattering features (broad features with many low intensity points), and label them in different clusters provided sufficient spatial separations, as shown in the next section.

Among many controls/tweaks one could apply in the clustering process in order to tailor DBSCAN to a certain need, we utilize its native ability of distinguishing in a cluster the "core" points from its boundary points [6]. This feature will play a critical role of intelligently reducing the size of the dataset and making it practical to interactively manipulate the 3D object created from the Turbine dataset.

## 3 Application on Neutron Data

Without loss of generality, we use 3D scatterplots for visualization after the DBSCAN clustering analysis. These scatterplots use a simple TF which maps the relative intensities of points in a cluster to a continuous alpha range in [0, 1], which makes a sharp contrast to the painstaking TF manipulation process demonstrated in Section 2.2.

### 3.1 Feature Extraction in Single Crystal Diffuse Scattering

Figure 4 shows the results from the DBSCAN clustering and the final visualization of the CZO dataset. Specifically, Figure 4(a) shows a 3D still image which includes all clusters identified by DBSCAN algorithm after using ε = 1.7 and weighted-minPts = 70. With this set of parameters, DBSCAN identifies 668 clusters in total, which covers 6.5% of the total ~45 million data points. In another word, 93.5% of the data points are identified as noise and will not be used for visualization. In comparison to traditional DVR result shown in Figure 3(b), DBSCAN has removed the cloudy background so efficiently that the spatially isolated features stand out, which is of tremendous help on visualizing morphological structures of the diffuse scattering patterns in the 3D reciprocal space.

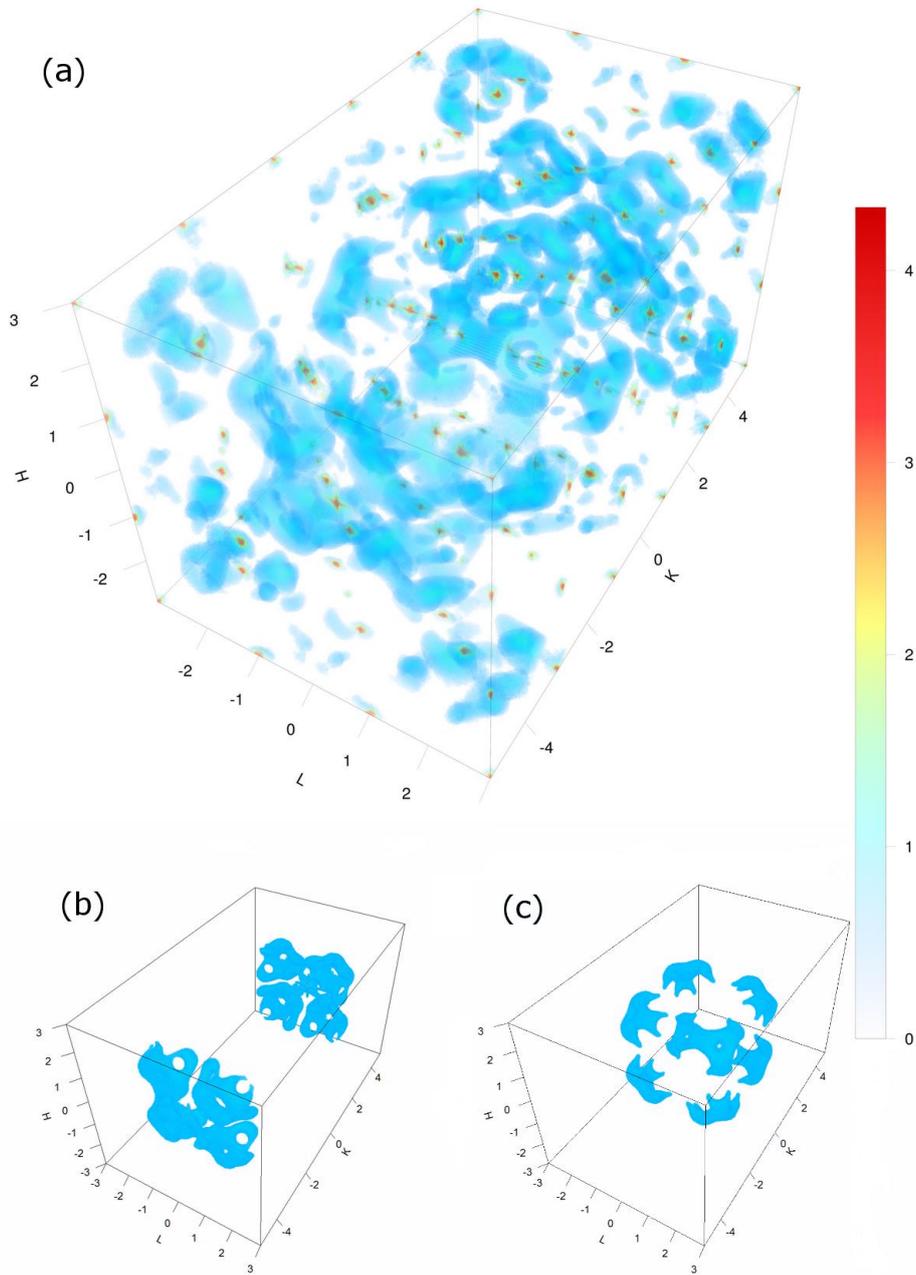

**Fig. 4.** (a) A 3D still image of the CZO dataset with all clusters identified by DBSCAN using ε = 1.7 and weighted-minPts = 70. (b, c) 3D scatterplots of clusters grouped per symmetry – (b) the most prominent two clusters; and (c) the next group of eight prominent clusters. Data are shown on the logarithmic scale of relative intensities.

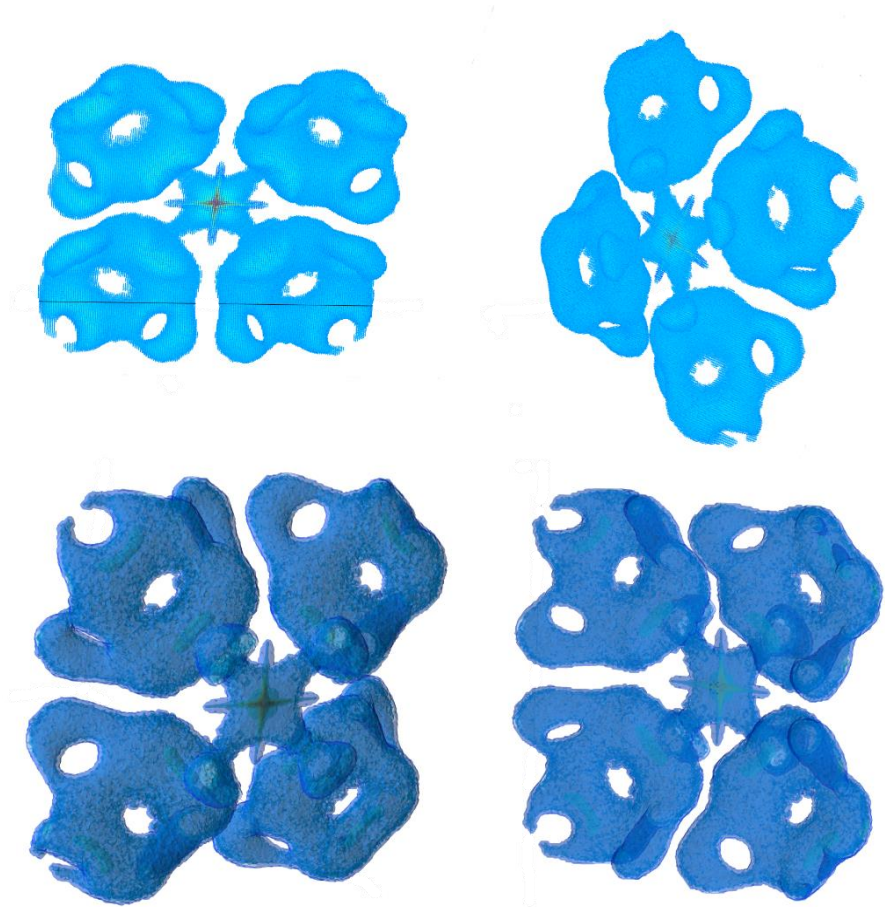

**Fig. 5.** Top panels - Scatterplots of one of the most prominent cluster identified in Figure 4(b) in different spatial perspectives; Bottom panels – iso-surfaces of the same object revealing finer internal structures.

More importantly, DBSCAN provides an easy way to extract distinct 3D diffuse scattering features from the volumetric data and make it possible to examine in detail each individual feature independently. For examples, Figure 4(b) and (c) show the first prominent group of two clusters and the second prominent group of eight clusters, respectively. Detailed close-ups could be easily achieved by simply selecting the desired clusters for visualization and such an example is given in Figure 5. The top two panels present scatterplots of the most prominent two clusters identified in Figure 4(b) in different spatial perspectives. The bottom two panels show the same object but plotted in its iso-surface format which reveal finer internal structures of the cluster.

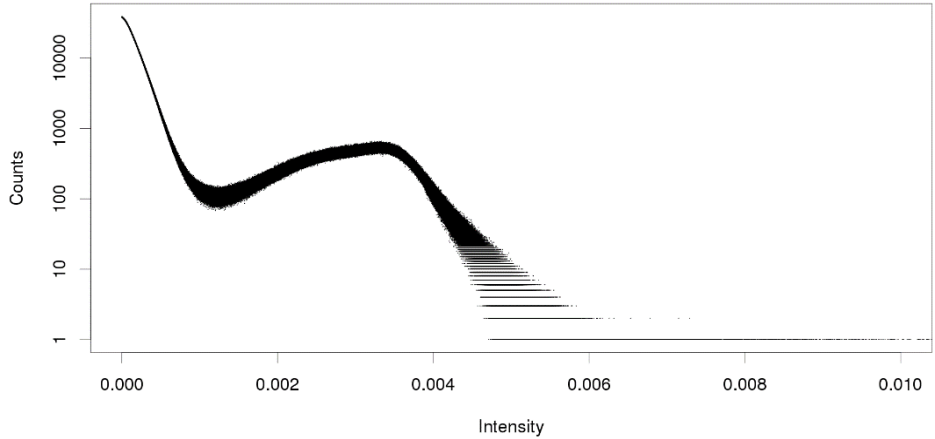

**Fig. 6.** Intensity histogram of the Turbine dataset. Negative intensity data points are not shown in the histogram.

### 3.2 Interactive Visualization of Neutron Tomography

Besides usual tasks like denoising, the neutron tomography dataset brings another challenge with its gigantic size to the visualization analysis. In our case, the raw Turbine dataset contains 8 billion data points in total and its intensity histogram is shown in Figure 6. Similar to the CZO dataset, the data points with low intensities ($<$ $10^{-3}$) are mostly from the background noise and can be filtered out without losing useful information (i.e. setting CUSP = $10^{-3}$). After applying this filter, the total data points to be fed into the analysis still remain a large number of ~340 millions.

Popular visualization packages (such as ParaView [13], Tomviz [14]) couldn't handle such large datasets easily in their original configuration. Although some professional products like VGStudio [15] can deal with this big data problem, it is impractical for neutron scientists and general facility users to use them daily for quick exploration of their datasets. Besides the high cost of the license and maintenance associated with these specialty software, the learning curve is usually so high that a dedicated staff member must devote a significant amount of time and effort to master the massive set of functionalities provided in these products.

To find a simple yet efficient way to deal with the big data problem in visualizing the tomography datasets, we found that the application of DBSCAN is easy to implement and can significantly improve the performance of the visualization, as demonstrated in rendering the 3D OpenGL object of the turbine blade interactively (see screenshots in Figure 7). The general procedure of applying DBSCAN to the tomography dataset is similar to that used in the feature extraction in Section 3.1 with one critical modification which is to apply the DBSCAN algorithm *twice* in the visualization analysis. The first DBSCAN identifies the bulk volume of the turbine blade as a single cluster and, at the same time, removes all the data points outside the bulk volume as noise. Comparing to traditional denoising procedure which sets a subjective intensity value as the cutting threshold and thus may risk feature losses, DBSCAN takes advantages of the spatial statistics of the imaged object to define its boundary. After the first DBSCAN clustering, the number of points representing the de-noised 3D object is reduced to ~110 millions and a 3D image is shown in Figure 7 (upper-left panel). However, it's still too big to create a smooth interactive visualization with such a large dataset. One way to solve this problem is to only render the surface of the object so that one could have a much smaller dataset for smoother interaction while using a moderately-equipped hardware. To get the surface data points, we perform a second DBSCAN clustering on the original dataset but explicitly demand it to drop all the boundary points so that the resulting cluster contains only the points confined within the bulk volume of the turbine blade. We then subtract all points included in this "peeled-off" cluster from those points in the cluster generated in the first DBSCAN. At the end, we get a "thinskin" (pure object-boundary) of the 3D turbine blade, as shown in Figure 7 (upper-right and two bottom panels). The final number of points to visualize is dramatically reduced to ~32 millions, or by a factor of 10 comparing to the original input. This double-DBSCAN approach raises the efficiency of the interactive visualization so much that the entire 3D object could be rendered on a personal laptop with a generic integrated graphics card.

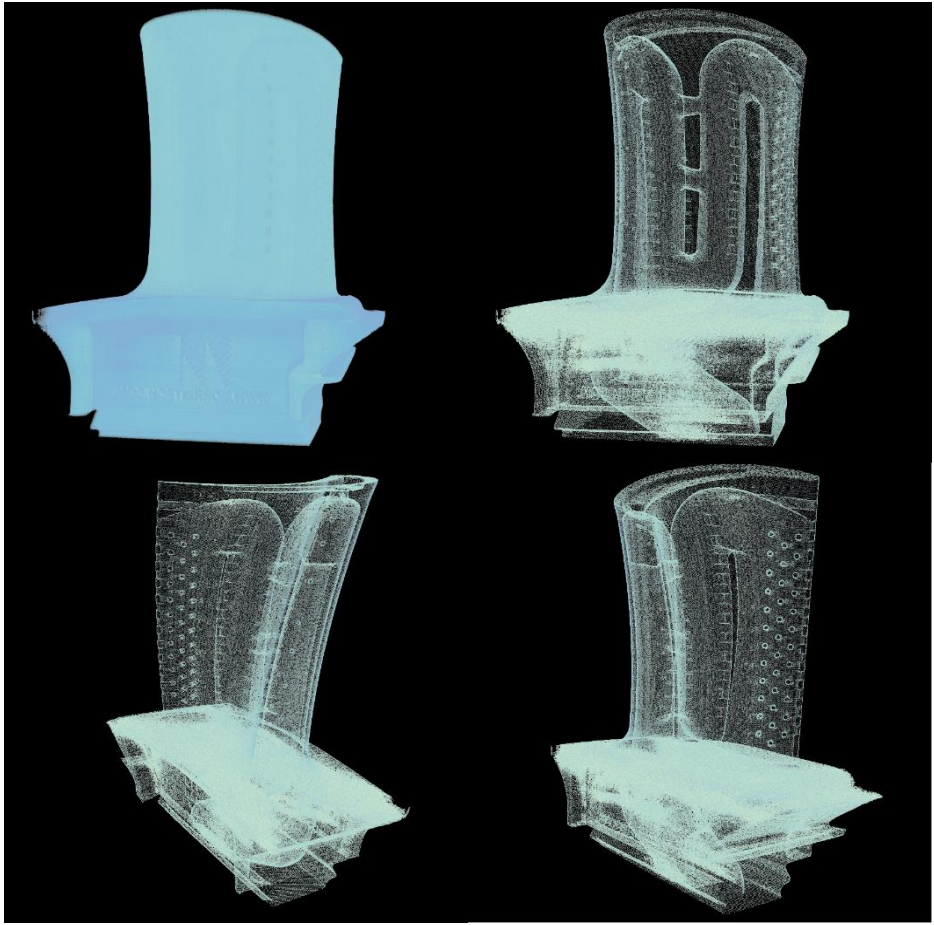

**Fig. 7.** Screenshots for (Upper-Left panel): the 3D image of the solid turbine blade after first-DBSCAN clustering; and (Upper-Right and two Bottom panels): a series of OpenGL renditions of the shell of the turbine blade after second-DBSCAN clustering, which shows the capability of manipulating this object in the interactive mode.

## 4    Discussion

### 4.1    Involvement of HPC

In our work, we have resorted to the computational resources provided by the Rhea cluster at Oak Ridge Leadership Computing Facility (OLCF) at ORNL. All still images presenting full datasets such as in Figure 3(b) and 4(a) are created on one of the Rhea's GPU nodes which provide two NVidia K80 GPUs with 1TB memory. Besides these tasks, other computationally intensive jobs such as the DBSCAN clustering on large datasets are also carried out on the Rhea nodes. As an initiative focusing on the viability of using DBSCAN to help on efficiently visualizing volumetric datasets, this work hasn't been exploring any potential performance gains from a distributed computing scheme. For example, all the DBSCAN clustering processes have been carried out with the R function provided in the "dbscan" package. Apparently, if replaced by a parallel implementation of the DBSCAN using MPI, the execution time of this algorithm (typically around 20 to 30 minutes for ~$10^8$ data points) will be much reduced (potentially linearly downscaled with requested computational resources).

Another call to distributed computing (which is the forte of the HPC platforms at OLCF) comes from the heavily burdened I/O system on the Rhea node. By using the HDF5 format to store the volumetric datasets used in this work, we have significantly reduced the READ_IN time (from disks to memory) by an order of magnitude (comparing to loading and combining tomography images in their TIFF format). However, it still takes 19 mins to read the 59 GB Turbine dataset into the memory. In future development, we could use the parallel HDF5 API in the I/O design to mitigate this performance bottleneck.

### 4.2    Expanding DBSCAN's Applications in Neutron Science

**Diffuse Scattering Research.** Analyzing volumetric single-crystal diffuse scattering data remains a high technique hurdle due to the large size of the datasets (typically a billion data points per dataset). The conventional approaches, such as 2D slicing, are tedious and error-prone to explore the vast volume in the 3D reciprocal space to identify characteristic features. Derived from our visualization analysis, intensity-weighted DBSCAN is found to be very effective in extracting distinguishable 3D features in these datasets. Given common geometrical characteristics existing in these features, a database of such features could be built upon the identification and curation of sufficiently ample examples. As a 3D visual library, the database could facilitate researchers to quickly get insight into underlying microstructural origins of the diffuse scattering features observed in their experiments. More importantly, it could potentially enable scientists to better correlate physical properties and local deviations from the average material structures (the latter of which is the cause of diffuse scattering), so that functional materials can be classified by their intrinsic local structures.

**Neutron Experiment Monitoring.** Further improvement on performance of our visualization analysis with the help of HPC resources at OLCF (discussed shortly in Section 4.1) could enable the analysis and visualization of real-time streaming data collected at multiple SNS/HFIR beamlines. By using such tools during experiments, users could identify weak scattering features at an early stage of the experiment and better plan the following steps so that the neutron beam time could be used more efficiently.

As a joint effort between OLCF and SNS, we have been working on the initial stage of integrating DBSCAN-aided visualization analysis into the Bellerophon Environment for Material Analysis (BEAM), a workflow management system developed at ORNL [16], which will connect the neutron data sources with the HPC at OLCF to speed up the analysis and visualization.

**Tomography Imaging.** As a visualization analysis approach, intensity-weighted DBSCAN could be generalized to applications on tomography imaging illuminated by not only neutron but also other particle sources (X-ray and electron, etc.). In this work, we have demonstrated the capability of this new approach with a simple example of interactive 3D rendition. This method could be extended to effectively detect 3D defects (e.g. cavities, cracks), 2D or 3D heterogeneous interfaces inside the imaged objects. All the analysis and visualization could be carried out onsite along with the ongoing imaging process.

## 5 Conclusion

We have investigated DBSCAN for neutron scattering data analysis and visualization with two examples, including feature extraction in a single crystal diffuse scattering dataset and interactive 3D visualization of neutron tomography dataset. In both cases, the intensity at each data point is of physical significance and can be used in DBSCAN clustering as the weight-factor to evaluate the input parameter minPts. By doing so, we are using both the photometric and spatial information of the scattering data in the visualization analysis, and it turns out to be a very effective approach to analyze and visualize large volumetric neutron datasets.